\documentclass{article}
\usepackage{makecell}
\usepackage{lineno}
\usepackage{bbm}
\usepackage{cite}
\usepackage{epsfig}
\usepackage{latexsym}
\usepackage{setspace}
\usepackage{flushend}
\usepackage{authblk}
\usepackage[margin=1in]{geometry}
\usepackage{float}
\usepackage{dsfont}
\usepackage{comment}
\usepackage{ upgreek }
\usepackage{caption}
\usepackage[backref,colorlinks,citecolor=blue,bookmarks=true]{hyperref}
\usepackage{stmaryrd}
\usepackage{xspace}
\usepackage{url}
\usepackage{amsmath,amssymb,amsfonts,amsthm,color,xspace}
\usepackage{subfigure,graphicx,tikz,enumitem}

\usepackage[utf8]{inputenc}
\usepackage[utf8]{inputenc} 
\usepackage[T1]{fontenc}    
\usepackage{hyperref}       
\usepackage{url}            
\usepackage{booktabs}       
\usepackage{amsfonts}       
\usepackage{nicefrac}       
\usepackage{microtype}      
\usepackage{comment}

\newtheorem{Theorem}{Theorem}
\newtheorem{Theorem*}{Theorem}

\newtheorem{Claim*}[Theorem]{Claim}

\newtheorem{CounterExample*}{$\overline{\hbox{\bf Example}}$}

\newtheorem{Example*}[Theorem]{Example}

\newtheorem{Intuition*}[Theorem]{Intuition}
\newtheorem{Joke*}[Theorem]{Joke}
\newtheorem{Lemma}[Theorem]{Lemma}

\newtheorem{Lemma*}[Theorem]{Lemma}
\newtheorem{Open problem}[Theorem]{Open problem}

\newtheorem{Question*}[Theorem]{Question}

\def \bSubexa    {\begin{subexa}}

\newcommand{\ignore}[1]{}

\newcommand{\RR}{\mathbb{R}}

\newcommand{\reals}{\RR}

\def \cB     {{\cal B}}

\def \cD     {{\cal D}}

\def \cM     {{\cal M}}

\def \cO     {{\cal O}}

\newcommand{\bi}{\begin{itemize}}
\newcommand{\ei}{\end{itemize}}

\def\orpro{\mathop{\mathchoice
   {\vee\kern-.49em\raise.7ex\hbox{$\cdot$}\kern.4em}
   {\vee\kern-.45em\raise.63ex\hbox{$\cdot$}\kern.2em}
   {\vee\kern-.4em\raise.3ex\hbox{$\cdot$}\kern.1em}
   {\vee\kern-.35em\raise2.2ex\hbox{$\cdot$}\kern.1em}}\limits}

\def\andpro{\mathop{\mathchoice
 {\wedge\kern-.46em\lower.69ex\hbox{$\cdot$}\kern.3em}
 {\wedge\kern-.46em\lower.58ex\hbox{$\cdot$}\kern.25em}
 {\wedge\kern-.38em\lower.5ex\hbox{$\cdot$}\kern.1em}
 {\wedge\kern-.3em\lower.5ex\hbox{$\cdot$}\kern.1em}}\limits}

\def\simge{\mathrel{%
   \rlap{\raise 0.511ex \hbox{$>$}}{\lower 0.511ex \hbox{$\sim$}}}}

\def\simle{\mathrel{
   \rlap{\raise 0.511ex \hbox{$<$}}{\lower 0.511ex \hbox{$\sim$}}}}

\newcommand{\tcr}[1]{\textcolor{red}{#1}}

\title{\fontsize{15.8}{19.8}\selectfont
Robust estimation algorithms don't need to know the corruption level.
}

\author{
\normalsize{Ayush Jain, Alon Orlitsky, Vaishakh Ravindrakumar}\\
  \normalsize{University of California, San Diego}\\
  \normalsize{\texttt{\{ayjain,alon,varavind\}@eng.ucsd.edu}}
}

\begin{document}
\maketitle
\noindent

\begin{abstract}
Real data are rarely pure.
Hence the past half century has seen great interest in robust estimation algorithms that perform well even when part of the data is corrupt. 
However, 
their vast majority approach optimal accuracy only when given a tight upper bound on the fraction of corrupt data.
Such bounds are not available in practice, resulting in weak guarantees and often poor performance. 
This brief note abstracts the complex and pervasive robustness problem into a simple geometric puzzle.
It then applies the puzzle's solution to derive a universal meta technique that converts any robust estimation algorithm requiring a tight corruption-level upper bound to achieve its optimal accuracy into one achieving essentially the same accuracy without using any upper bounds.
\looseness-1
\end{abstract}

\section{Motivation}
Much of statistics concerns learning from samples, typically assumed to be generated by an unknown distribution.
As the number of samples increases, many statistical tasks can be learned to arbitrarily small error. 
However, in practical applications, often not all samples are generated according to the underlying distribution. 
Some may be inaccurate, 
corrupt, or even adversarial.

\ignore{
Robust algorithms that learn accurately even when some samples are corrupt, were therefore studied since the early works of Tukey~\cite{Tukey60}, Huber~\cite{Huber64}, and has been the subject of many a book,~\cite{Huber09,HampelRRS11}
\tcr{, surveys~\cite{} and research articles~\cite{}.} (remove, can't find)
The research continues to this day~\cite{surveys}\tcr{survey, remove?}, with recent efficient algorithms for robust mean and density estimation~\cite{LaiRV16,DiakonikolasKKLMS16,DiakonikolasKKLMS17}, sparse mean estimation~\cite{BalakrishnanDLS17,DiakonikolasKKPS19}, learning from batches~\cite{QiaoV18,ChenLM20a,JainO20a,JainO20b,ChenLM20b,JainO21c}, Erdos-Renyi graphs~\cite{AcharyaJKSZ}, and many more~\cite{particular_works}.
}

Robust algorithms that learn accurately even when some samples are corrupt, were therefore studied since the early works of Tukey~\cite{Tukey60}, Huber~\cite{Huber64}, and has been the subject of many a book,~\cite{Huber09,HampelRRS11}.
The research continues to this day with recent efficient algorithms for robust mean and density estimation~\cite{LaiRV16,DiakonikolasKKLMS16,DiakonikolasKKLMS17}, sparse mean estimation~\cite{BalakrishnanDLS17,DiakonikolasKKPS19}, learning from batches~\cite{QiaoV18,ChenLM20a,JainO20a,JainO20b,ChenLM20b,JainO21c}, Erdos-Renyi graphs~\cite{AcharyaJKSZ21}, and many more~\cite{BalakrishnanDLS17,CharikarSV17,KlivansKM18,SteinhardtCV18,DiakonikolasKKLMS18,HopkinsL18,KothariSS18,DiakonikolasKKLSS19,DiakonikolasKKLMS19,ZhuJS19,PrasadSBR20a,LiuSLC20,ChenKMY20,PensiaJL21,JambulapatiLTT21}, see~\cite{DiakonikolasK19} for a recent survey.

Essentially all these works consider the \emph{Huber model}~\cite{Huber64} and its generalizations. 
A fraction $1-\alpha$ of the samples are genuine, while the remaining $\alpha$ fraction may be inaccurate, corrupt, or adversarial, even chosen after viewing the genuine samples.
A significant effort within this research concerns \emph{statistical estimation}, approximating an underlying distribution or parameter set. 
This note addresses statistical estimation under any of the corruption models above.

Most robust algorithms, both in general and specifically for estimation, 
either make the theoretically convenient assumption that $\alpha$ is known, or more practically utilize an input parameter $\beta$ that is assumed to upper bound $\alpha$, and 
else may result in an arbitrary error. 
Since $\alpha$ is never known, the first approach cannot work in practice.
Assuming an upper bound necessitates a large $\beta$ to ensure correctness.
Yet as the following examples show, the algorithm's estimation error is an increasing function $f(\beta)$.

\cite{DiakonikolasKKLMS16,DiakonikolasKKLMS17} considered robust learning of high-dimensional 
Gaussians with identity covariance. 
They showed that as long as the corrupted fraction $\alpha$ is at most the upper bound $\beta$, with sufficiently many samples, the distribution mean can be leaned in $\ell_2$ distance $\cO(\beta\sqrt{\log(1/\beta)})$ and the same accuracy applies for learning the distribution itself in TV-Distance. 
They also derived similar results for learning general Gaussian distributions and product distributions when provided an upper bound $\beta\ge\alpha$. 
A sequence of papers,~\cite{QiaoV18,ChenLM20a,JainO20a,JainO20b,ChenLM20b,JainO21c} studied discrete and piecewise-polynomial continuous distributions where the samples arrive in batches of size $n$, and a fraction $\alpha\le\beta$ of the batches can be arbitrary.
They showed that the underlying distribution can be learned to a TV distance $\cO(\beta\sqrt{\log(1/\beta)/n})$.
Similarly,~\cite{AcharyaJKSZ21} showed that for Erd\H{o}s-R\'enyi graphs with $n$ nodes, where the edges connected to a fraction $\alpha\le\beta$ of the nodes may be corrupt, the connection probability $p$ can be learned to accuracy $\cO(\beta\sqrt{\frac{p(1-p)\log(1/\beta)}{n}}+  \frac\beta n\log n+\frac{\sqrt{p(1-p)\log n}}{n})$.

While this note addresses robust estimation, the known-upper-bound assumption is prevalent in many other robust learning applications including robust regression and robust gradient descent, see for example~\cite{KlivansKM18,ChenKMY20,DiakonikolasKS19,PrasadSBR20a}.
It would interesting to see whether they could be similarly addressed as well.

The conflicting dependence of accuracy and validity on $\beta$ raises several natural concerns. 
First the corrupt fraction $\alpha$ is typically unknown.
Upper bounding it by a fixed $\beta$ goes against the very essence of robustness. 
Even if one is willing to assume an upper bound $\beta\ge \alpha$, what should it be.
Large $\beta$ can drastically reduce the algorithm's accuracy, for example,
yet small $\beta$ risks invalidating its results altogether. 

Questions about the validity and choice of $\beta$ have therefore haunted this approach in both presentations and applications.

This brief note takes a bird's-eye view of optimal robust estimation.
Instead of addressing each individual problem, it reformulates all of them as an elementary geometric puzzle whose exceedingly simple and elegant solution
yields a universal, unified, and efficient method achieving optimal  error without knowing or bounding $\alpha$.\looseness-1

The next section describes the puzzle and its solution,  
Section~\ref{sec:app} applies the result to remove the upper bound requirement from all robust estimation problems, and achieving the same accuracy as if the corruption level was known in advance. 
It demonstrates the effect on the three robust learning examples provided above. 
Section~\ref{sec:limitations} considers some of the technique's limitations and possible extensions.


\section{AirTag}\label{sec:airtag}
Apple's AirTag approximates the location $x$ of a misplaced item.
Its beta-version 
successor lets the user select an approximation distance $\beta$ that in turn  affects the search space $S(\beta)$, if $\beta'\le\beta$ then $S(\beta')\subseteq S(\beta)$.
AirTag then returns an approximate location $x_\beta$ that is within distance $\beta$ from $x$ if $x\in S(\beta)$, and is arbitrary otherwise.
The set of possible locations and distance may form any metric space, $\beta$ can assume any finite set of positive values, and par for Apple's course, except for its growth with $\beta$, $S(\beta)$ is completely unknown.

The best approximation distance is clearly the smallest $\beta$ such that $x\in S(\beta)$, which we denote by $\alpha$. 
Choosing $\beta>\alpha$ worsens $x_\beta$'s accuracy, while for $\beta<\alpha$, $x_\beta$ is arbitrary. 
However $x$ and $S(\beta)$ are unknown, hence so is $\alpha$. 
Upon obtaining the locations $x_\beta$ for all $\beta$, can you approximate $x$ to a distance at most $c\cdot\alpha$ for some small constant $c$?

We will soon describe two simple solutions, but first observe that 
the puzzle captures both the essence and functionality of robust algorithms. 
Given an upper bound $\beta$ on the corruption level, these algorithms approximate an underlying distribution or parameter set. 
If the actual corruption level is below $\beta$, their output is within the specified distance from the distribution, and otherwise, no guarantees are provided. 

One small difference is that for estimation algorithms, the distance guarantee is not $\beta$, but rather some known increasing function $f(\beta)$ specific to each problem. 
The addition of $f$ can be viewed as a simple reparamtrization of $\beta$, hence it does not change the problem. 
Yet it will be convenient to use it in the application, hence we phrase the results in this more general form.

Consider a metric space $(\cM,d)$, finite set $\cB\subseteq \reals$, 
an association $x_\beta\in \cM$ with every $\beta\in \cB$, and a non-decreasing $f:\cB\to[0,\infty)$.
An \emph{estimator} is given $x_\beta$ for all $\beta\in \cB$ as input, and outputs a point $\hat x\in \cM$. 
It achieves \emph{approximation factor} $c$ if for every input and every $x\in\cM$ and $\alpha\in \cB$ such that $d(x,x_{\beta})\le f(\beta)$ for all $\beta\ge\alpha$, 
it must satisfy
$d(\hat x,x)\le c\cdot f(\alpha)$.

The following estimator achieves approximation factor 2.
Let $B(x,r):=\{y\in \cM:d(x,y)\le r\}$ be the ball of radius $r\ge 0$ around $x\in\cM$.
Define $\beta'$ to be the smallest number in $\cB$ such that $\bigcap_{ \cB\ni\beta\ge \beta'} B(x_\beta, f(\beta))$ is non empty, and let $x'$ be any point in this intersection.

\begin{Lemma}\label{lem:2}
$x'$ achieves approximation factor 2.
\end{Lemma}
\begin{proof}
Let $x\in\cM$ and $\alpha\in \cB$ satisfy $d(x,x_{\beta})\le f(\beta)$ for all $\beta\ge\alpha$.
For all $\beta\ge \alpha$, $x\in B(x_\beta, f(\beta))$. Hence, by definition $\beta'\le \alpha$, and  $x'\in B(x_\alpha,f(\alpha))$.
By the triangle inequality $d(x,x')\le d(x,x_\alpha)+d(x_\alpha,x')\le 2f(\alpha)$.
\end{proof}

The next lemma shows that approximation factor 2 is best possible.
\begin{Lemma}\label{lem:lb}
No estimator achieves approximation factor less than 2.
\end{Lemma}
\begin{proof}
Consider the {reals} with absolute difference distance, $\cB=\{\epsilon,1\}$ for
$\epsilon\in(0,0.5]$, $f(\beta)=\beta$ for all $\beta$, and $x_\epsilon=1+\epsilon$, $x_1=0$.
Let estimator $\hat x$ achieve approximation factor $c$.
For every $x\in\reals$ and $\alpha\in \cB$ such that $|x-x_{\beta}|\le f(\beta)$ for all $\beta\ge\alpha$, it must satisfy $|\hat x-x|\le c\cdot f(\alpha)$.
Hence for $x=1$ and $\alpha=\epsilon$, 
$|\hat x-1|\le c\epsilon $, 
while for $x=-1$ and $\alpha=1$,
$|\hat x-(-1)|\le c$.
By the triangle inequality,
$c+c\epsilon \ge |\hat x-1|+|\hat x+1|\ge 2$, hence $c\ge 2/(1+\epsilon)$ that can be made to exceed any number less than 2.
\end{proof}


While finding an $x'$ at the intersection of several balls may be feasible in some metric spaces, for example when $\cM=\reals$, for general $(\cM,d)$ this may be computationally challenging.
The next method achieves a slightly larger approximation factor $c=3$, but requires only pairwise distance computations among the $x_{\beta}$'s. 

Define $\hat\beta$ to be the smallest number in $\cB$ such that for all $\cB\ni\beta\ge\hat\beta$, $d(x_\beta, x_{\hat\beta})\le f(\beta)+f(\hat\beta)$.


\begin{Lemma}\label{lem:3}
$x_{\hat\beta}$ achieves approximation factor 3.
\end{Lemma}
\begin{proof}
Let $x\in\cM$ and $\alpha\in \cB$ satisfy $d(x,x_{\beta})\le f(\beta)$ for all $\beta\ge\alpha$. By the triangle inequality, for all $\beta\ge \alpha$,
\[
d(x_{\beta}, x_{\alpha})
\le
d(x_{\beta}, x)+d(x, x_{\alpha})
\le 
f(\beta)+f({\alpha}),
\]
hence by $\hat \beta$'s definition, $\hat\beta\le \alpha$.
Adding the triangle inequality, the condition on $\alpha$, and $\alpha\ge \hat \beta$,
\[
d(x_{\hat\beta},x)
\le
d(x_{\hat\beta},x_{\alpha})+d(x_{\alpha},x)
\le
f(\hat\beta)+f(\alpha) + f(\alpha)
\le
3f(\alpha).
\qedhere
\]
\end{proof}
Returning to the AirTag, Lemmas~\ref{lem:2} and~\ref{lem:3} provides estimators that locate the item to distance $2\alpha$ and $3\alpha$.

\section{Robustness applications}\label{sec:app}

The AirTag puzzle and its solution suggest a simple universal method for removing the upper bound requirement from any robust estimation algorithm. 

Recall that many existing algorithms utilize an input parameter $\beta$, and so long as it exceeds the actual fraction of corrupt data $\alpha$, they estimate the unknown distribution or parameter vector $x$ to error $f(\beta)$, while for $\beta<\alpha$ the algorithm fails and its error can be arbitrary large.
If $\alpha$ was known in advance, one could run the algorithm with input parameter $\beta=\alpha$ resulting in output $x_{\alpha}$ that achieves the best error guarantee $d(x,x_{\alpha})\le f(\alpha)$.
We show that even without knowing $\alpha$, we can still find an estimate $\hat x\in \cM$ such that $d(x,\hat x)\le 3\cdot f(\alpha)$. 

Let $\beta_{\max}$ denote the algorithm's \emph{breakdown point}, the largest corruption fraction for which the algorithm gives a meaningful answer.
For example, when more than half the data is corrupt, no parameter can be accurately estimated, hence for every algorithm, $\beta_{\max}<1/2$.
We henceforth assume that the actual corruption level $\alpha\le\beta_{\max}$.

The AirTag solution involves finding $x_\beta$ potentially for every $\beta$ in the set $\cB$ of all possible $\alpha$ values.
In corruption applications, every $\alpha\in[0,\beta_{\max}]$ is possible, hence the approach cannot be applied directly. 
Instead, we select a small geometric sequence $B\subseteq [0,\beta_{\max}]$ that contains a tight approximation of any $\alpha\in[0,\beta_{\max}]$.
For any choice of $\theta>1$ and $\epsilon >0$, let
$\cB$ be the collection of $\beta_i:=\beta_{\max}/\theta^i$ for $i=0,1,\ldots,$ till $\beta_i\le f^{-1}(\epsilon)$.
Let $\alpha':=\min\{\cB\ni\beta\ge \alpha\}$ be the closest upper bound of $\alpha$ in $B$.
Clearly $\alpha'\le \max\{\theta \alpha, f^{-1}(\epsilon)\}$, and 
since $\alpha'\ge \alpha$, for all $\beta\ge \alpha'$ we have $d(x,x_\beta)\le f(\beta)$.\looseness-1

Running the original algorithm $|\cB| \le \lceil\log_\theta (1/ f^{-1}(\epsilon)\rceil$ times, once for each
value in $\cB$, Lemmas~\ref{lem:2} and~\ref{lem:3} achieve error $c\cdot f(\alpha')\le c\cdot\max\{f(\theta\alpha),\epsilon\}$ for $c=2$ and $3$, respectively.
For example, selecting $\theta=1.1$, the method achieves 
$\le c\cdot\max\{f(1.1\alpha),\epsilon\}$ error by running the algorithm $\cO(\log(1/f^{-1}(\epsilon))$ times. 
For typical problems $f(\beta)$ is sublinear in $\beta$, hence the method achieves $\le c\cdot\max\{1.1f(\alpha),\epsilon\}$ error.

This approach applies to every robust estimation algorithm.
It converts any robust estimation algorithm that requires a tight upper bound on $\alpha$ to achieve its optimal guarantee, into one that achieves essentially the same guarantees without knowing $\alpha$.
Consider for example the problems mentioned in the introduction. 
For any positive $\epsilon$, however small, without knowledge or upper bound on the corruption level $\alpha$, 
high-dimensional identity-covariance Gaussians can be learned to TV distance $\cO(\alpha\sqrt{\log(1/\alpha)})+\epsilon$ and the same holds for approximating the mean in $\ell_2$ distance.
Similar results hold for the other problems considered in \cite{DiakonikolasKKLMS16,DiakonikolasKKLMS17}. 
When the samples arrive in batches of $n$, the underlying distribution can be estimated to TV distance $\cO(\alpha\sqrt{\log(1/\alpha)/n})+\epsilon$.


For Erd\H{o}s-R\'enyi graphs, recall that for any $\beta\in [\alpha,\beta_{\max}]$, one can estimate the connection probability $p$ to accuracy $f(\beta,p):= \cO(\beta\sqrt{\frac{p(1-p)\log(1/\beta)}{n}}+  \frac\beta n\log n+\frac{\sqrt{p(1-p)\log n}}{n})$.
However, since $p$ is unknown, so is this accuracy, hence we cannot apply our approach directly. 
To mitigate this problem, we first use the worst possible corruption level $\beta_{\max}$ to approximate $p$ by a weak approximation $\tilde p$, and then us $\tilde p$ to obtain a tight upper bound on $f(\beta,p)$ for all $\beta$, which therefore upper bounds the actual error all $\beta\ge \alpha$.

A simple calculation shows that for all $0\le p,\tilde p\le1$, $\varepsilon\ge 0$, and $\beta$, 
if $\tilde p(1-\tilde p)-p(1-p)\in[0,\varepsilon]$ then 
$f(\beta,\tilde p)-f(\beta,p)\in[0,\cO(\beta\sqrt{\frac{\varepsilon\log(1/\beta)}n}+\frac{\sqrt{\varepsilon\log n}}n)]$.
Therefore, given such $\tilde p$, $f(\beta,\tilde p)$ upper bounds the algorithm's for all input parameter $\beta\ge \alpha$, using our approach we can estimate $p$ to an accuracy $c\cdot f(\alpha,\tilde p)= c\cdot f(\alpha,p) +  \cO(\alpha\sqrt{\frac{\varepsilon\log(1/\alpha)}n}+\frac{\sqrt{\varepsilon\log n}}n)+\epsilon$.

Next, we obtain a weak estimate of $p$ by using the max corruption level as the input parameter, namely, $\beta= \beta_{\max}$. This way we can estimate $p$ (and hence $p(1-p)$) to an accuracy $f(\beta_{\max},p)$, and using $\beta_{\max} = \cO(1)$ we get $f(\beta_{\max},p)=\cO(\sqrt{\frac{p(1-p)}{n}}+  \frac{\log n}n)\le \cO(p(1-p)+  \frac{\log n}n)$.
This estimate of $p$, can be used to find $\tilde p$ such that  $\tilde p(1-\tilde p)-p(1-p)\in[0,\varepsilon]$, where $\varepsilon = \cO(p(1-p)+  \frac{\log n}n)$. Using this $\tilde p$, as shown above, we can estimate $p$ to an accuracy $c\cdot f(\alpha,\tilde p) =  c\cdot f(\alpha,p) +  \cO(\alpha\sqrt{\frac{\varepsilon\log(1/\alpha)}n}+\frac{\sqrt{\varepsilon\log n}}n)+\epsilon= c\cdot f(\alpha,p) +  \cO(
\frac{\log n}{n^{3/2}})+\epsilon$. Note that the extra $\frac{\log n}{n^{3/2}}$ term is very small and dominated by $f(\alpha,p)$ except for the extreme regime where $p(1-p)=\cO( \frac{\log n}{n})$ and $\alpha=\cO( \frac{1}{\sqrt{n}})$.

\section{Extensions and limitations}\label{sec:limitations}

This note concerns robust estimation of an underlying distribution, or some of its parameters, from samples.
It would be interesting to generalize this approach to other robust learning problems such as regression, gradient descent etc.
In fact, as the AirTag puzzle itself suggests, the approach is not limited to robustness problems.
It applies to any algorithm whose accuracy depends monotonically on an input parameter, up to an unknown limit where the algorithm is no longer correct. 
It would be interesting to find other types of problems with that property. 

Another natural generalization may be to multi parameter problems.
Yet, even with two parameters we can not achieve guarantees similar to the one parameter case.
Consider a metric space $(\cM,d)$, a finite set $\cB$ and for 
let $f({\beta_1,\beta_2})$ for $\beta_1,\beta_2\in \cB$ be non-negative and non-decreasing function in both $\beta_1$ and $\beta_2$.
For each $\beta,\gamma\in \cB$, let $x_{\beta,\gamma}$ be points in $\cM$.
Can we find 
a point $\hat x\in \cM$,  
such that for all $x\in \cM$ and $\alpha_1,\alpha_2\in \cB$ if $\forall\, \beta_1\ge \alpha_1$ and $\beta_2\ge \alpha_2$, $d(x_{\beta_1,\beta_2},x)\le f({\beta_1,\beta_2})$, then the point $\hat x$ satisfies $d(\hat x, x)\le c\cdot f({\alpha_1,\alpha_2})$ for some universal constant $c>0$?

Unfortunately such a point does not necessarily exist.
Consider the reals with absolute difference distance and $\cB=\{0,1\}$.
Let $x_{1,1}=0$, $x_{0,1}=1$, $x_{1,0}=-1$, and $x_{1,1}$ be arbitrary.
Let $f({1,1})=1$ and $f({\beta_1,\beta_2})=0$ for $(\beta_1,\beta_2)\neq (1,1)$.
For $ x = 1$, $(\alpha_1,\alpha_2)= (0,1)$ satisfies the above condition, and $f(\alpha_1,\alpha_2)=0$.
For $x = -1$, $(\alpha_1,\alpha_2)= (1,0)$ satisfies the above condition, and $f(\alpha_1,\alpha_2)=0$.
Then for $\hat x $ to be within a distance constant times $f(\alpha_1,\alpha_2)$ from all corresponding $x$, the point $\hat x$ must have a distance zero from both 1 and -1, therefore $\hat x$ does not exist.\looseness-1

In fact, this limitation is not an artifact of our approach, but is inherent to some 2-parameter estimation problems. 

Let $\cD$ be an unknown distribution over $\reals$ with unknown mean $x$ and variance $\sigma^2$.
Given a tight upper bound on either $\sigma$ or $\alpha$ simple truncated-mean estimators achieve optimal error $\Theta(\sigma\sqrt{\alpha})$.
When a tight upper bound on $\alpha$ is known, trim $\alpha$ fraction of samples from both the extremes,
and 
when a tight upper bound on $\sigma$ is known, 
recursively remove the farthest sample from the mean of remaining samples until the mean square deviation of the remaining samples is $\cO(\sigma^2)$.
We show that if both $\alpha$ and $\sigma$ can be arbitrary, then for any choice of $\epsilon$ and $C$ any algorithm 
incurs an error much larger than $C\sigma\sqrt{\alpha}+\epsilon$ for some $\alpha$ and $\sigma$.

Consider two distributions, $\cD_1$ assigns probability $1$ to $0$, while $\cD_2$ assigns probability $9/10$ to $0$ and probability $1/10$ to $21\epsilon$.
For $\alpha>1/10$ and $\cD=\cD_1$ the adversary can make the distribution of overall samples appear to be $\cD_2$. In this case $\sigma=0$, therefore for any estimate $\hat x$ that achieve the error at most $C\sigma\sqrt{\alpha}+\epsilon$ then must satisfy $\hat x\le \epsilon$.
However, for the case $\alpha=0$, $\cD=\cD_2$, where the mean of $\cD$ is $21\epsilon/10$, for any $\hat x\le \epsilon$, the error $|2.1\epsilon-\hat x|\ge 1.1\epsilon> C\sigma\sqrt{\alpha}+\epsilon$.

\bibliographystyle{alpha}
\bibliography{biblio}

\end{document}